\documentclass[final,5p,times,twocolumn]{elsarticle}

\usepackage{xcolor} % 用于 \textcolor
\usepackage{amsmath} % 一般推荐加载数学包，虽然这里可能不需要

\usepackage{graphicx}  % 用于 \resizebox
\usepackage{booktabs}  % 用于 \cmidrule 等表格命令
\usepackage{amssymb}
\usepackage{amsmath,amsfonts}
\usepackage{algorithmic}
\usepackage{array}
\usepackage[caption=false,font=normalsize,labelfont=sf,textfont=sf]{subfig}
\usepackage{textcomp}
\usepackage{stfloats}
\usepackage{url}
\usepackage{verbatim}
\usepackage{balance}
\usepackage{chngcntr}
\usepackage{amssymb, amsmath}
\usepackage{bm}
\usepackage{color}
\usepackage{threeparttable}
\usepackage{amsmath}
\usepackage{color}
\usepackage{pifont}
\usepackage{array} 
\usepackage{multirow}
\usepackage{tabularx}
\usepackage{algorithm}
\usepackage{algorithmic}
\usepackage{tabularray}
\usepackage{makecell}
\usepackage[section]{placeins}
\usepackage{wrapfig}
\usepackage{graphicx}

\journal{Nuclear Physics B}

\begin{document}

\begin{frontmatter}



\title{DiffRIS: Enhancing Referring Remote Sensing Image Segmentation with Pre-trained Text-to-Image Diffusion Models}


\author[1]{Zhe Dong} %% Author name
\author[1]{Yu-Zhe Sun} %% Author name
\author[1]{Tian-Zhu Liu} %% Author name
\author[1]{Yan-Feng Gu\corref{cor1}} %% Author name
\ead{guyf@hit.edu.cn} % 通讯作者的邮箱
\cortext[cor1]{Corresponding author.} % 通讯作者脚注内容
\affiliation[1]{organization={School of Electronics and Information Engineering, Harbin Institute of Technology},%Department and Organization
            city={Harbin},
            postcode={150001}, 
            country={China}}

%% Abstract
\begin{abstract}
Referring remote sensing image segmentation (RRSIS) enables the precise delineation of regions within remote sensing imagery through natural language descriptions, serving critical applications in disaster response, urban development, and environmental monitoring. Despite recent advances, current approaches face significant challenges in processing aerial imagery due to complex object characteristics including scale variations, diverse orientations, and semantic ambiguities inherent to the overhead perspective. To address these limitations, we propose DiffRIS, a novel framework that harnesses the semantic understanding capabilities of pre-trained text-to-image diffusion models for enhanced cross-modal alignment in RRSIS tasks. Our framework introduces two key innovations: a context perception adapter (CP-adapter) that dynamically refines linguistic features through global context modeling and object-aware reasoning, and a progressive cross-modal reasoning decoder (PCMRD) that iteratively aligns textual descriptions with visual regions for precise segmentation. The CP-adapter bridges the domain gap between general vision-language understanding and remote sensing applications, while PCMRD enables fine-grained semantic alignment through multi-scale feature interaction. Comprehensive experiments on three benchmark datasets—RRSIS-D, RefSegRS, and RISBench—demonstrate that DiffRIS consistently outperforms existing methods across all standard metrics, establishing a new state-of-the-art for RRSIS tasks. The significant performance improvements validate the effectiveness of leveraging pre-trained diffusion models for remote sensing applications through our proposed adaptive framework.
\end{abstract}

%% Keywords

\end{frontmatter}




\section{Introduction}

Referring remote sensing image segmentation (RRSIS) aims to identify specific regions in remote sensing imagery based on given textual conditions, making it particularly suitable for practical applications such as defense reconnaissance\cite{kaplan2022monitoring}, climate impact studies\cite{rolnick2022tackling}, urban infrastructure management\cite{huang2022city3d}, and land use categorization\cite{avtar2020assessing}. Unlike traditional single-modal segmentation methods\cite{dong2023distilling, dong2023spatial}, RRSIS leverages textual descriptions to guide image segmentation, overcoming the limitations of fixed category labels and enabling the processing of more diverse vocabulary and syntactic variations. However, the spatial and geographical disparities conveyed from an aerial perspective are fundamentally distinct from those in natural images, presenting significant challenges in achieving accurate and contextually relevant visual segmentation. Therefore, effectively harnessing textual conditions to enhance segmentation precision and semantic consistency remains a critical challenge in RRSIS research.

In recent years, multimodal fusion methods have made significant strides in modeling the semantic relationships between images and natural language for RRSIS tasks. Various approaches have been developed to enhance the complementarity of visual and linguistic information, including attention mechanisms \cite{luo2020cascade, hu2020bi, yang2021bottom}, multi-level feature fusion \cite{zhang2022coupalign, chen2022position}, auto-regressive vertex generation \cite{zhu2022seqtr, liu2023polyformer}, and expression queries \cite{luo2020multi, huang2020referring, ding2021vision}. Attention mechanisms effectively establish correspondences between visual and textual data, facilitating more accurate semantic fusion and improving segmentation performance by associating image pixels or regions with relevant words in the text. Multi-level feature fusion methods, on the other hand, integrate image features and textual information across multiple scales, enhancing the model’s sensitivity to small targets and intricate details, while also reducing computational complexity and improving efficiency in real-time, high-precision scenarios. Auto-regressive vertex generation empowers the decoder to directly produce coordinate sequences in the coordinate space, thereby mitigating encoding redundancy and uncertainty, while leveraging sequential information to guide the order and positioning of vertices for more accurate and complete segmentation. Lastly, expression query methods optimize segmentation by retrieving target regions based on varied queries, each reflecting a unique interpretation of the expression, which further refines target differentiation and recognition accuracy. However, despite these advancements, existing methods face notable limitations, including challenges in generalization to diverse and complex datasets, inefficiencies in handling ambiguously defined regions, and reliance on task-specific designs or predefined queries, which restrict their adaptability and robustness in addressing the inherent variability and complexity of RRSIS tasks.

Text-to-image diffusion models\cite{zhang2023adding,saharia2022photorealistic} have recently demonstrated remarkable capabilities in the field of generative modeling, particularly excelling in capturing intricate semantics and fine-grained visual details. By leveraging pre-training on large-scale image-text datasets (e.g., LAION-5B\cite{schuhmann2022laion}), these models effectively learn and integrate both high-level semantic concepts and low-level visual attributes, enabling highly controlled generation through customizable textual prompts. Their exceptional scalability allows for the production of high-quality images characterized by rich textures, diverse content, and coherent structures, while also exhibiting flexibility in semantic composition and editing. Moreover, the latent visual features learned by text-to-image diffusion models show a strong correlation with the corresponding words in textual prompts. This implicit alignment capability further enhances their performance in cross-modal semantic understanding and generation, making them a powerful tool for multimodal applications.

Building on the aforementioned advantages, the integration of text-to-image diffusion models into RRSIS tasks holds significant promise. On the one hand, text-to-image diffusion models, through extensive pre-training on vision-language datasets, effectively capture the intricate semantic associations between natural language descriptions and target regions within remote sensing imagery. By achieving deep cross-modal alignment, diffusion models enable a more precise mapping of textual descriptions to corresponding regions in remote sensing images, thereby offering robust guidance for referring image segmentation in complex and heterogeneous scenarios. On the other hand, the latent feature representations generated by diffusion models encapsulate not only rich visual details but also comprehensive global contextual information, providing essential semantic support for identifying multi-scale and fine-grained targets within remote sensing imagery. Given the variability in target scales, ranging from minute structures to vast expanses, diffusion models excel by unifying low-level and high-level visual concepts during generation. This unified modeling approach significantly enhances the sensitivity to multi-scale targets, thereby addressing the unique challenges posed by remote sensing data.

To this end, we explore the application of text-to-image diffusion models in remote sensing scenarios and devise DiffRIS, an effective diffusion model tailored for referring remote sensing image segmentation (RRSIS). By leveraging the powerful pre-trained knowledge embedded in text-to-image diffusion models, DiffRIS bridges the gap between natural language descriptions and remote sensing images, enabling precise segmentation of complex and diverse regions. Besides, a context-perception adapter (CP-adapter) and a progressive cross-modal reasoning decoder (PCMRD) are incorporated into the proposed framework, with both designed to enhance cross-modal semantic alignment and improve fine-grained segmentation accuracy. Extensive evaluations on multiple benchmark datasets demonstrate that DiffRIS outperforms existing methods, setting a new benchmark for RRSIS tasks.

In summary, the main contributions of this study are summarized as follows.

\begin{itemize}

	\item[(1)] We pioneer the first diffusion model-based framework for RRSIS tasks. By leveraging the rich, pre-trained knowledge embedded in text-to-image diffusion models, DiffRIS significantly improves cross-modal semantic alignment and boosts segmentation accuracy.
	
	\item[(2)] Two complementary modules are designed: the context-perception adapter (CP-adapter), which refines textual features by capturing contextual dependencies, and the progressive cross-modal reasoning decoder (PCMRD), which iteratively refines semantic alignment to ensure precise segmentation.
	
	\item[(3)] Extensive evaluations across three benchmark datasets demonstrate that DiffRIS achieves state-of-the-art performance, surpassing existing methods and establishing a new standard for RRSIS tasks in terms of both precision and robustness.
	
\end{itemize}

The remainder of the paper is organized as follows: Section~\ref{section:Related Work} reviews related works on RRSIS. In Section~\ref{section:methods}, we describe the proposed methodology in detail. Section~\ref{section:EXPERIMENTS} presents a comprehensive set of experiments and in-depth analyses. Finally, Section~\ref{section:Conclusion} concludes the paper and offers insights into potential future research directions.

\section{Related Work}
\label{section:Related Work}

\subsection{Diffusion Models in Remote Sensing}

The diffusion models, also referred to as the diffusion probabilistic models, have emerged as a powerful family of deep generative models. Essentially, it consists of a set of probabilistic generative models that systematically degrade data by injecting noise, subsequently learning to reverse this process to generate samples. Recently, diffusion models have spurred significant advancements in the field of remote sensing research, demonstrating a sustained and substantial impact.

Diffusion models have the capacity to synthesize realistic remote sensing images from existing images or given textual descriptions, thereby facilitating the advancement of various remote sensing applications. Ou \textit{et al.}\cite{ou2023method} initially employed a vision-language pre-training model to generate captions for remote sensing images, thereby obtaining preliminary text prompts. Subsequently, they utilized these text prompts to enable the Stable Diffusion\cite{rombach2022high} model to synthesize the desired remote sensing images. Geolocation and sampling time were employed as prompts in the Stable Diffusion model by Khanna \textit{et al.}\cite{khanna2023diffusionsat}, effectively enhancing its ability to generate high-quality satellite images. Moreover, by utilizing relevant features of remote sensing images as control conditions, Tang \textit{et al.}\cite{tang2024crs} refined the generative process of the Stable Diffusion model. Recent studies have focused on employing masks such as class labels\cite{wu2023high}, maps\cite{espinosa2023generate}, and semantic layouts\cite{yuan2023efficient, zhao2023label, baghirli2023satdm} as guiding images for diffusion models to achieve image-to-image generation.

Another application of diffusion models in the field of remote sensing involves image enhancement, such as super-resolution, cloud removal, and denoising. Shi \textit{et al.}\cite{shi2023hyperspectral} utilized cascaded images of multispectral and hyperspectral data as conditional inputs for the diffusion model, leveraging valuable information captured from both modalities to generate high spatial-resolution hyperspectral images (HSIs). By employing text prompts and edge information\cite{xie2015holistically} as guiding conditions, Czerkawski \textit{et al.}\cite{czerkawski2024exploring} integrated cloud masks and diffuse cloud-free images to achieve cloud removal. Yu \textit{et al.}\cite{yu2024universal} enhanced the diffusion model's ability to mitigate system noise by simulating adverse imaging conditions for remote sensing satellites through the introduction of various attack disturbances in the input images.

\subsection{Referring Image Segmentation}

Referring image segmentation (RIS) is a complex multimodal task that necessitates effective coordination between language and vision for accurate target region segmentation, surpassing the challenges posed by visual question answering\cite{antol2015vqa} and visual dialogue\cite{kottur2018visual}. 

Early approaches utilized long short-term memory (LSTM)\cite{greff2016lstm} for encoding linguistic representations while employing convolutional neural network (CNN)\cite{he2016deep} to extract spatial features from images at multiple levels. The recursive multimodal interaction model (RMI)\cite{liu2017recurrent} employed a multimodal LSTM network for linguistic representation, integrating visual features and capturing spatial variations of multimodal information to generate coarse localization masks, which were subsequently refined through a unidirectional LSTM network. Hu \textit{et al.}\cite{hu2016segmentation} achieved referring image segmentation by employing a CNN-LSTM framework to extract visual features from images and linguistic features from natural language representations. A recurrent refinement network (RRN)\cite{li2018referring} was proposed to address the lack of multi-scale semantics in image representation, utilizing a feature pyramid to match each word with every pixel in the image to generate an initial segmentation mask, which is subsequently refined and iteratively optimized through a recursive optimization module.

Recent developments have increasingly favored the adoption of Transformers to enhance the feature extraction and integration of visual-language modalities. The language-aware visual Transformer (LAVT)\cite{yang2022lavt} employed the Swin Transformer\cite{liu2021swin} as the visual backbone, integrating vision-language fusion modules within the final layers of the visual encoder. Ding \textit{et al.} developed a vision-language Transformer (VLT)\cite{ding2022vlt} framework to facilitate deep interactions among multimodal information, thereby enhancing the comprehensive understanding of vision-language features. CRIS\cite{wang2022cris} and ReSTR\cite{kim2022restr} employed analogous methodologies, utilizing dual transformer encoders for the preliminary encoding of modalities, which was subsequently succeeded by feature integration via a multi-modal Transformer encoder or decoder. Similarly, PolyFormer\cite{liu2023polyformer} and SeqTr\cite{zhu2022seqtr} utilized a multi-modal Transformer for vision-language fusion, producing masks as sequences of contour points. In contrast, CGFormer\cite{tang2023contrastive} and GRES\cite{liu2023gres} treated Transformer queries as region proposals, framing RIS as proposal-level classification tasks.

\subsection{Referring Remote Sensing Image Segmentation}

As an emerging research direction within the field of remote sensing, RRSIS offers a user-friendly approach for individuals lacking specialized knowledge in this domain to specify objects of interest and conduct targeted image analysis. Building upon LAVT, Yuan \textit{et al.}\cite{yuan2024rrsis} proposed the first RRSIS framework. They also introduced a language-guided cross-scale enhancement (LGCE) module that effectively integrated deep and shallow features by leveraging linguistic cues, thereby improving segmentation performance for small and spatially dispersed objects within remote sensing images. Additionally, a language-guided cross-scale enhancement (LGCE) module was designed to integrate deep and shallow features by leveraging linguistic cues, enhancing segmentation performance for small and spatially dispersed objects within remote sensing images. To address the challenges posed by the diverse spatial scales and orientations present in aerial imagery, the rotated multi-scale interaction network (RMSIN)\cite{liu2024rotated} incorporated several pivotal components aimed at enhancing model performance. First, the intra-scale interaction module (IIM) was introduced to effectively capture fine-grained details across multiple scales. Subsequently, the cross-scale interaction module (CIM) was designed to integrate these details cohesively, ensuring consistency and coherence throughout the network. Additionally, RMSIN employed an adaptive rotated convolution (ARC) to adeptly accommodate the varied orientations of objects, thereby significantly improving segmentation accuracy. In addition, a cross-modal bidirectional interaction model (CroBIM)\cite{dong2024cross} was proposed to explore the interaction and alignment of visual and linguistic features across multiple levels. CroBIM incorporated a context-aware prompt modulation (CAPM) module to enhance text feature encoding with multi-scale visual context, thereby enabling effective spatial understanding of target objects. Furthermore, the language-guided feature aggregation (LGFA) module fostered interaction between multi-scale visual representations and linguistic features, capturing cross-scale dependencies and refining performance through an attention deficit compensation mechanism. Finally, the mutual-interaction decoder (MID) facilitated precise vision-language alignment via cascaded bidirectional cross-attention, resulting in highly accurate segmentation masks and marking a significant advancement in RRSIS tasks.

\section{Methodology}
\label{section:methods}

\begin{figure}[tbp]
	\begin{center}
		\centerline{\includegraphics[width=1\linewidth]{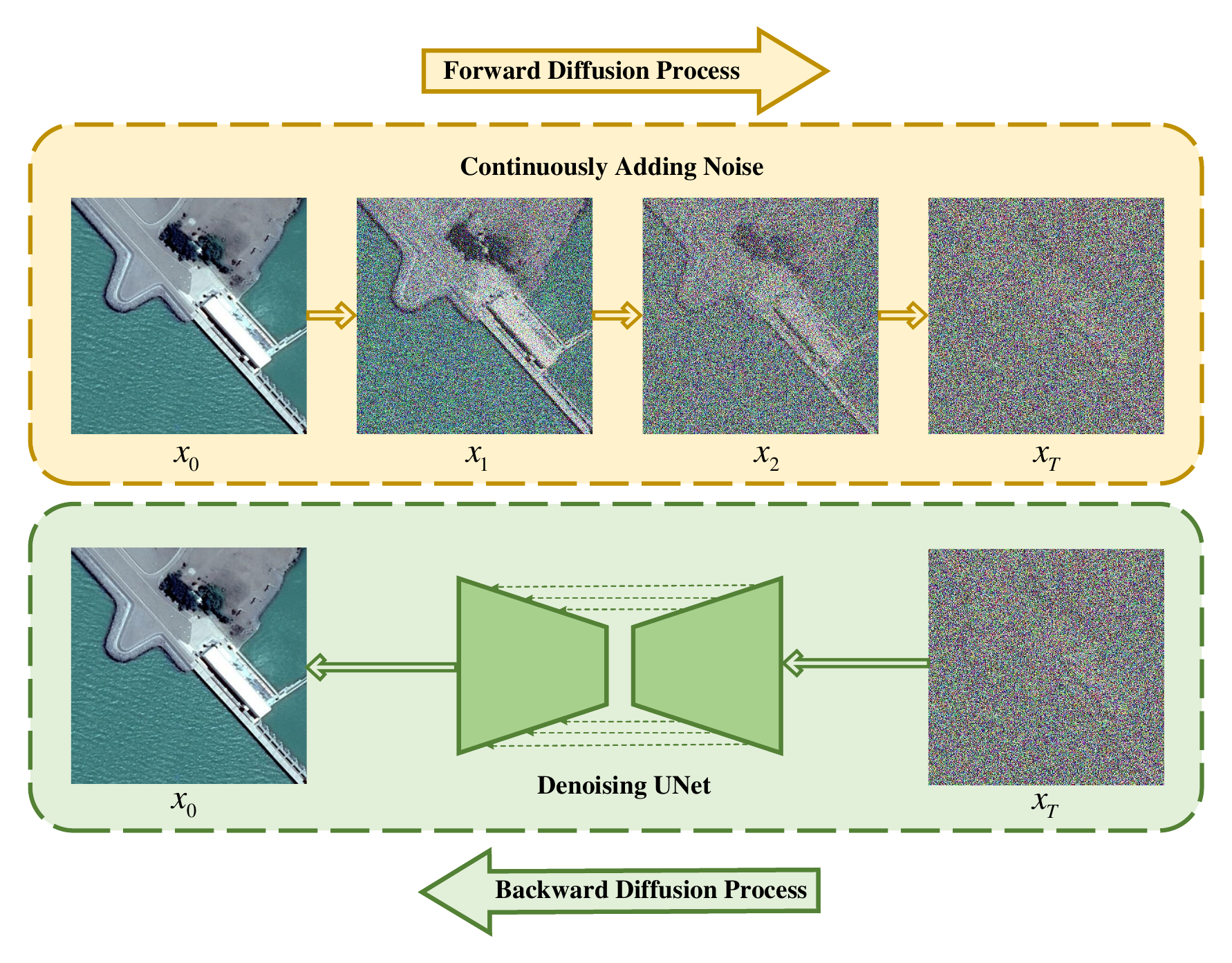}}
		\caption{Schematic illustration of diffusion probabilistic models. The forward diffusion process (top) gradually adds noise to the original image. The backward diffusion process (bottom) learns to reverse this degradation through a denoising UNet, progressively recovering the original image from pure noise.}\label{diffusion}
	\end{center}
\end{figure}

\subsection{Preliminaries: Diffusion Models}
\label{section:submethod1}

Diffusion models, also known as diffusion probabilistic models, fundamentally rely on a parameterized Markov chain trained via variational inference. The primary objective of diffusion models is to facilitate sample generation by preserving a process of data perturbation through noise, referred to as diffusion. This approach aims to transform a random distribution (i.e., noise) into a probability distribution that aligns with the distribution of the observed dataset, thereby enabling the generation of desired outcomes through sampling from this probability distribution. The conceptual foundations of diffusion models can be traced back to 2015\cite{sohl2015deep}; however, their popularity surged following the publication of denoising diffusion probabilistic models (DDPM)\cite{ho2020denoising} in 2020, subsequently igniting exponential growth in research interest within the computer vision (CV) field.

\textbf{Forward Diffusion Process:} As shown in Fig.~\ref{diffusion}, given the original image $x_0$, the forward diffusion process incrementally adds Gaussian noise through $T$ iterations, resulting in a sequence of progressively noise-perturbed images $x_1,x_2,\cdots,x_T$ until the images are entirely degraded\cite{ho2020denoising, yang2023diffusion, zhang2023text}. Throughout this process, each generated image $x_t$ is solely dependent on its preceding image $x_{t-1}$. Consequently, this procedure can be effectively modeled as a Markov chain:

\begin{figure*}[tbp]
	\begin{center}
		\centerline{\includegraphics[width=1\linewidth]{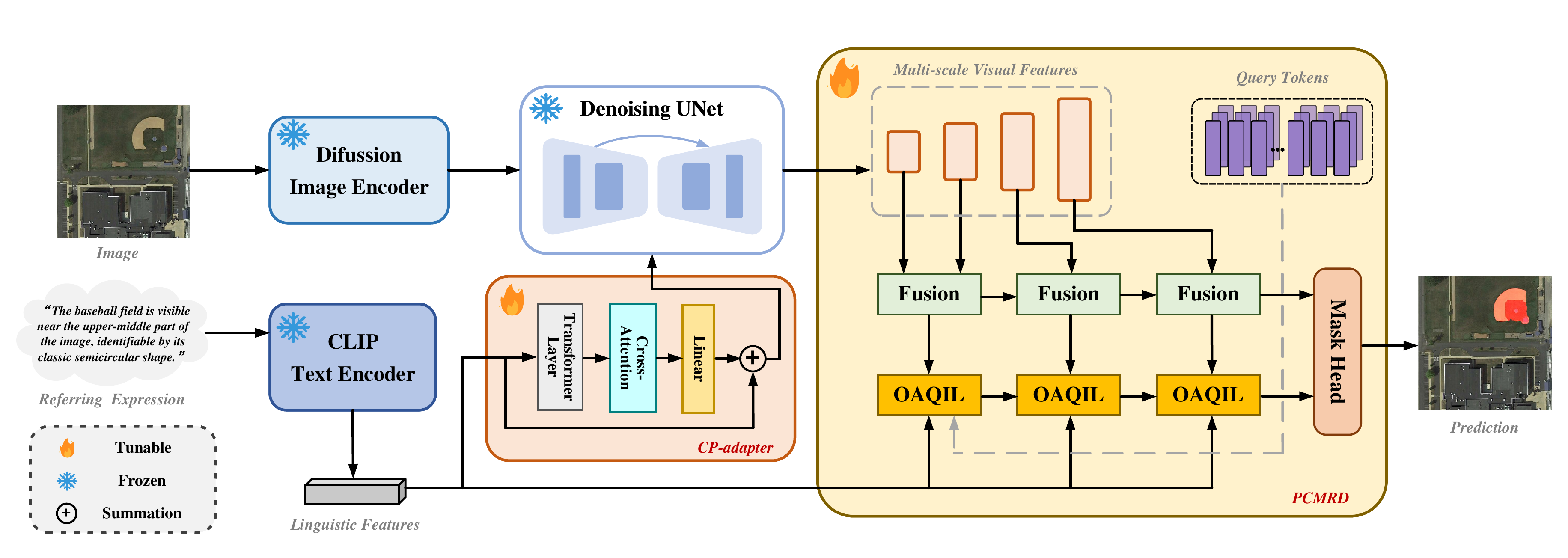}}
		\caption{Architecture of the proposed DiffRIS framework for referring remote sensing image segmentation. The framework consists of three main components: (1) pre-trained CLIP text encoder, diffusion image encoder and denoising UNet that collaboratively extract multi-modal features, (2) a context perception adapter (CP-adapter) that refines linguistic features and bridges the domain gap, and (3) a progressive cross-modal reasoning decoder (PCMRD) that enables fine-grained segmentation through cross-modal interaction. During training, only the CP-adapter and PCMRD are optimized while all pre-trained components remain frozen to preserve their learned representations.}\label{framework}
	\end{center}
\end{figure*}

\begin{equation}
	q\left(x_t \vert x_{t-1}\right)=\mathcal{N}\left(x_t ; \sqrt{1-\beta_t} x_{t-1}, \beta_t I\right)
\end{equation}

\begin{equation}
	\begin{aligned}
		q\left(x_{1: T} \vert x_0\right) & =\prod_{t=1}^T q\left(x_t \vert x_{t-1}\right) \\
		& =\prod_{t=1}^T \mathcal{N}\left(x_t ; \sqrt{1-\beta_{t-1}}, \beta_t I\right)
	\end{aligned}
\end{equation}where $q\left(x_{1: T} \vert x_0\right)$ represents the transition probability of the Markov chain, characterizing the distribution of Gaussian noise added at each step. The parameter $\beta$ serves as a hyperparameter for the variance of the Gaussian distribution , increasing linearly with $t$. Additionally, $I$ denotes the identity matrix with dimentions matching those of the input image $x_0$.

For simplicity, we assume that Gaussian noise acts as the transition kernels. Specifically, using the notation $\alpha_t=1-\beta_t$ and $\bar{\alpha}_t=\prod_{i=1}^t \alpha_i$, it is theoretically feasible to obtain a noisy image at any step , as illustrated below:

\begin{equation}
	q\left(x_t \vert x_0\right)=\mathcal{N}\left(x_t ; \sqrt{\bar{\alpha}_t} x_0,\left(1-\bar{\alpha}_t\right) I\right)
\end{equation}Consequently, as $T$ approaches infinity, $x_T$ is capable of converging to the standard normal distribution $\mathcal{N}(0, I)$.

\textbf{Backward Diffusion Process:} The backward diffusion process aims to learn the reverse transition probability $q(x_{t-1} \vert x_t)$, thereby gradually recovering the image from noise:

\begin{equation}
	p_\theta\left(x_{t-1} \vert x_t\right)=\mathcal{N}\left(x_{t-1} ; \mu_\theta\left(x_t, t\right), \Sigma_\theta\left(x_t, t\right)\right)
\end{equation}where $\theta$ denotes the parameters of the UNet\cite{ronneberger2015u} architecture that are subject to optimization. Consequently, the backward diffusion process can be articulated as follows:
\begin{equation}
	p_\theta\left(x_{0: T}\right)=p\left(x_T\right) \prod_{t=1}^T p_\theta\left(x_{t-1} \vert x_t\right)
\end{equation}

The training objective of the network is to minimize the Kullback-Leibler (KL) divergence, thereby ensuring that the backward  diffusion process $p_\theta\left(x_0, x_1, \ldots, x_T\right)$ aligns with the forward diffusion process $q\left(x_0, x_1, \ldots, x_T\right)$:

\begin{equation}
	\mathcal{L}(\theta)=\operatorname{KL}\left(q\left(x_0, x_1, \ldots, x_T\right) \| p_\theta\left(x_0, x_1, \ldots, x_T\right)\right)
\end{equation}

\subsection{DiffRIS Framework}

The proposed DiffRIS framework is designed to fully leverage the capabilities of text-to-image diffusion models by transferring their pre-trained knowledge from large-scale vision-language datasets to remote sensing applications, specifically for RRSIS tasks. By seamlessly integrating high-level semantic representations with low-level visual attributes, DiffRIS achieves robust cross-modal semantic alignment, effectively mapping textual descriptions to their corresponding regions within remote sensing imagery. With the incorporation of a query mechanism and explicit cross-modal guidance, DiffRIS excels in capturing multi-scale, fine-grained targets in complex, heterogeneous environments. Moreover, it harnesses the global contextual information inherent in generative models, thereby enhancing semantic understanding and improving performance in challenging remote sensing scenarios.

The overall architecture of the proposed DiffRIS is illustrated in Fig.~\ref{framework}. The input image $I \in \mathbb{R}^{H \times W \times 3}$ is initially transformed into a latent space representation $\textbf{z} \in \mathbb{R}^{h \times w \times c}$ by a pre-trained diffusion image encoder (i.e., VQGAN\cite{esser2021taming}), where $H$ and $W$ are the height and width of the image, and $h, w, c$ denote the dimensions of the latent representation. Simultaneously, a referring language expression $E=\left\{e_i\right\}, i \in \{0, \ldots, N\}$ , where $N$ is the number of tokens, is encoded into linguistic features $L \in \mathbb{R}^{l_m \times D_l}$ using a CLIP-based\cite{radford2021learning} text encoder. In this context, $l_m$ specifies the maximum length of the token sequence, while $D_l$ indicates the dimensionality of the linguistic feature space. 

To mitigate the domain gap between the pre-training task and downstream tasks, we introduce a context perception adapter (CP-adapter) to obtain the refined linguistic features $L^{\prime}$. The refined linguistic features $L^{\prime}$ and the latent image representation $\textbf{z}$ are then fed into a UNet-based denoising network to extracts multi-scale visual features $\left\{V_i \in \mathbb{R}^{H_i \times W_i \times C_i}\right\}_{i=1}^4$, where $\left(H_i, W_i\right)=\left(H / 2^{i+1}, W / 2^{i+1}\right)$ and $C_i$ denote the spatial resolution and channel dimension of the $i$-th visual feature, respectively. Unlike traditional diffusion processes, no noise is added to the latent representation, and the denoising step is simplified to a single forward pass, streamlining feature extraction for downstream tasks. 
Furthermore, we propose a progressive cross-modal reasoning decoder (PCMRD), which incorporates learnable query tokens $Q$ to represent objects. By alternately querying linguistic features $L$ and grouping multi-scale visual features $\left\{V_i \right\}_{i=1}^4$ into these tokens, the PCMRD enables object-aware interactions, thereby facilitating cross-modal reasoning to produce the final predictions. 

During the training phase of the proposed DiffRIS framework, only the parameters of the proposed CP-adapter and PCMRD are updated, while all other parameters remain frozen. This strategy ensures that the pre-trained knowledge of the diffusion model and the vision-language encoders is preserved, allowing the framework to efficiently adapt to remote sensing scenarios.

\subsection{Context Perception Adapter}

To effectively preserve the pre-trained knowledge of the text encoder and bridge the domain gap between pre-training tasks and downstream applications, we propose a context perception adapter (CP-adapter). The CP-adapter is specifically designed to refine linguistic features, enabling them to adapt to the domain-specific contextual requirements inherent in RRSIS tasks. This adapter utilizes a lightweight yet efficient architecture, incorporating three key components: global context modeling, object-aware reasoning, and domain-specific adjustment, ensuring optimal feature alignment and task adaptability.

Initially, the Transformer encoder is employed to capture the global semantic relationships among linguistic tokens. The input linguistic features $L$ are recursively processed through multiple layers of the Transformer encoder, resulting in preliminary context-aware representations $\mathbf{H}$:

\begin{figure}[tbp]
	\begin{center}
		\centerline{\includegraphics[width=0.8\linewidth]{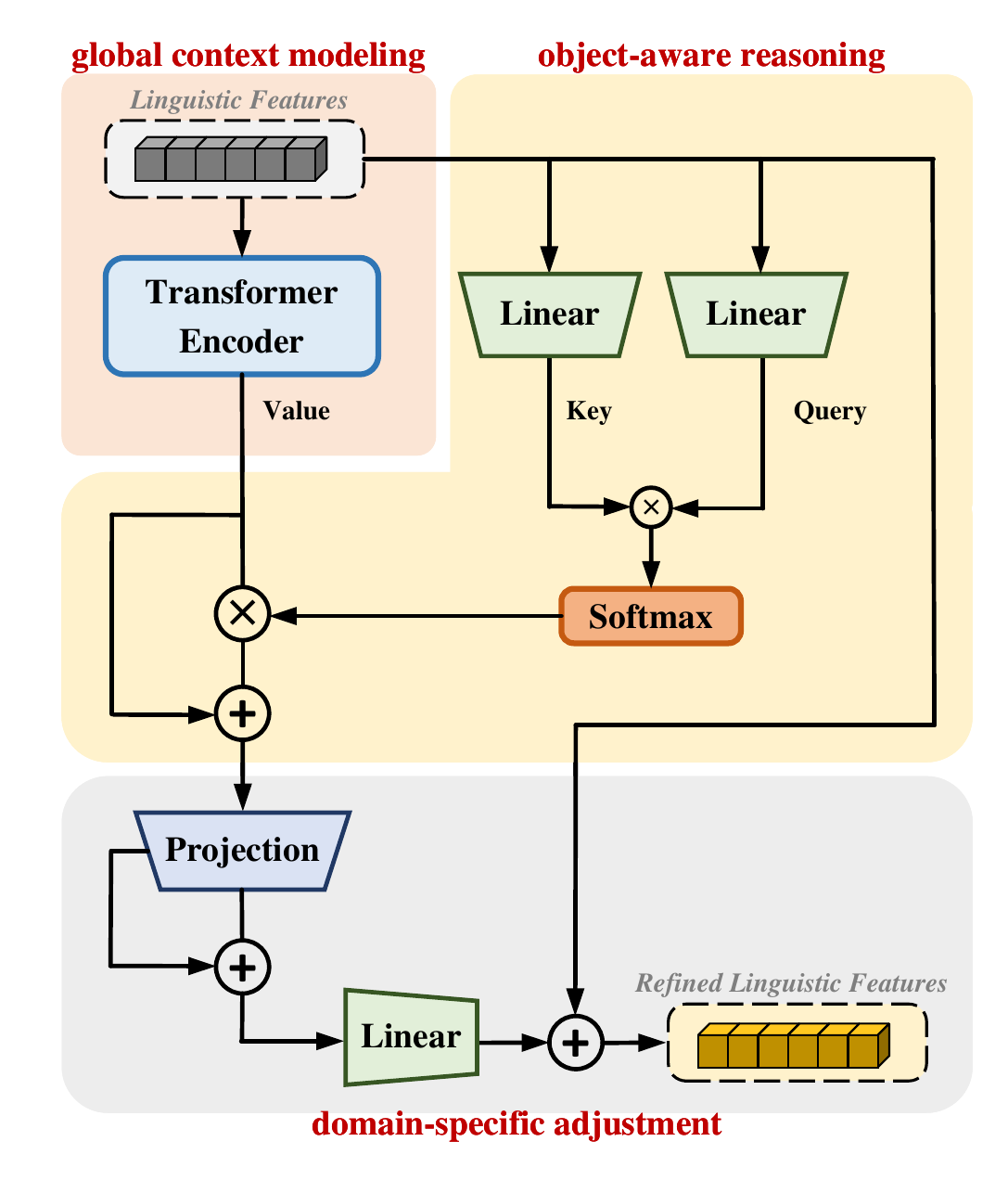}}
		\setlength{\belowcaptionskip}{0.2pt}  % 标题与下方内容之间的空白
		\caption{Detailed architecture of the proposed  CP-adapter. The adapter comprises three key components: global context modeling via a Transformer encoder, object-aware reasoning through attention mechanisms, and domain-specific adjustment using low-rank adaptation. The refined linguistic features preserve pre-trained knowledge while adapting to remote sensing domain characteristics.}\label{cp-adapter}
	\end{center}
\end{figure}

\begin{equation}
	\mathbf{H}=\text { TransformerEncoder }(L)\in \mathbb{R}^{l_m \times D_l},
\end{equation}In each layer of the Transformer encoder, global semantic dependencies are captured through a combination of multi-head self-attention (MHSA) mechanisms and feedforward neural networks.

Subsequently, task-relevant information is enhanced by iteratively applying attention mechanisms between the encoded features and the original input. Specifically, the query ($Q$), key ($K$), and value ($V$) are generated as follows:

\begin{equation}
	\mathbf{H}_v=\mathbf{W}_v  \mathbf{H}, \quad L_k=\mathbf{W}_k  L, \quad L_q=\mathbf{W}_q  L,
\end{equation}where $ \mathbf{W}_v, \mathbf{W}_k, \mathbf{W}_q \in \mathbb{R}^{D_l \times D_l}$ are linear projection matrices. The attention score is then calculated as:

\begin{equation}
	\operatorname{Attention}(L_q, l_k, \mathbf{H}_v)=\operatorname{softmax}\left(\frac{L_q L_k^{\top}}{\sqrt{D_l}}\right) \mathbf{H}_v,
\end{equation}

Subsequently, it is integrated with the input linguistic features $L$ to generate enhanced feature representations $\mathbf{H}_{\text{Enhanced}}$. 

\begin{equation}
	\mathbf{H}^\prime=\operatorname{Attention}(L_q, l_k, \mathbf{H}_v)+L,
\end{equation}
\begin{equation}
	\mathbf{H}_{\text {Enhanced }}=\operatorname{Projection}(\mathbf{H}^\prime) + \mathbf{H}^\prime,
\end{equation}where $\operatorname{Projection}(\cdot)$ refers to a sequential block of two fully connected layers with a ReLU activation in between.

To further align the features with the requirements of downstream tasks, domain-specific adjustments are applied through a low-rank adaptation layer:

\begin{equation}
	\mathbf{H}_{\text {LLM }}=\mathbf{W}_{\text {LLM }} \cdot \mathbf{H}_{\text {Enhanced }},
\end{equation}where $\mathbf{W}_{\text {LLM }}$ represents the low-rank linear transformation matrix, with initial weights set to zero to prevent excessive perturbation  during fine-tuning.

The CP-adapter generates the refined linguistic features $\hat{L}$ by weighted fusion of the original features $L$ and the adjusted features $\mathbf{H}_{\mathrm{LLM}}$:
\begin{equation}
	\hat{L}=\alpha \cdot \mathbf{H}_{\mathrm{LLM}}+(1-\alpha) \cdot L,
\end{equation} where $\alpha$ is a learnable weight that balances the contribution of the adapted features and the original features.

Through the collaborative effect of the aforementioned steps, the CP-adapter not only preserves the rich semantic information encoded during the pre-training phase but also adapts to the specific requirements of downstream tasks. This effectively bridges the domain gap while maintaining high computational efficiency.

\subsection{Progressive Cross-Modal Reasoning Decoder}

Furthermore, we propose a progressive cross-modal reasoning decoder (PCMRD) that incorporates learnable query tokens for object representation. Through alternately performing query-language interaction and query-vision feature grouping across multiple scales, PCMRD enables fine-grained cross-modal reasoning, thereby enhancing the model's capability in comprehending textual descriptions and localizing corresponding visual targets. PCMRD comprises three progressive decoding processes, each of which begins by integrating two levels of visual feature maps. These fused features are further combined with linguistic features and query tokens to serve as the input to the object-aware query interaction layer (OAQIL), which subsequently outputs the updated tokens.

\begin{figure}[tbp]
	\begin{center}
		\centerline{\includegraphics[width=0.7\linewidth]{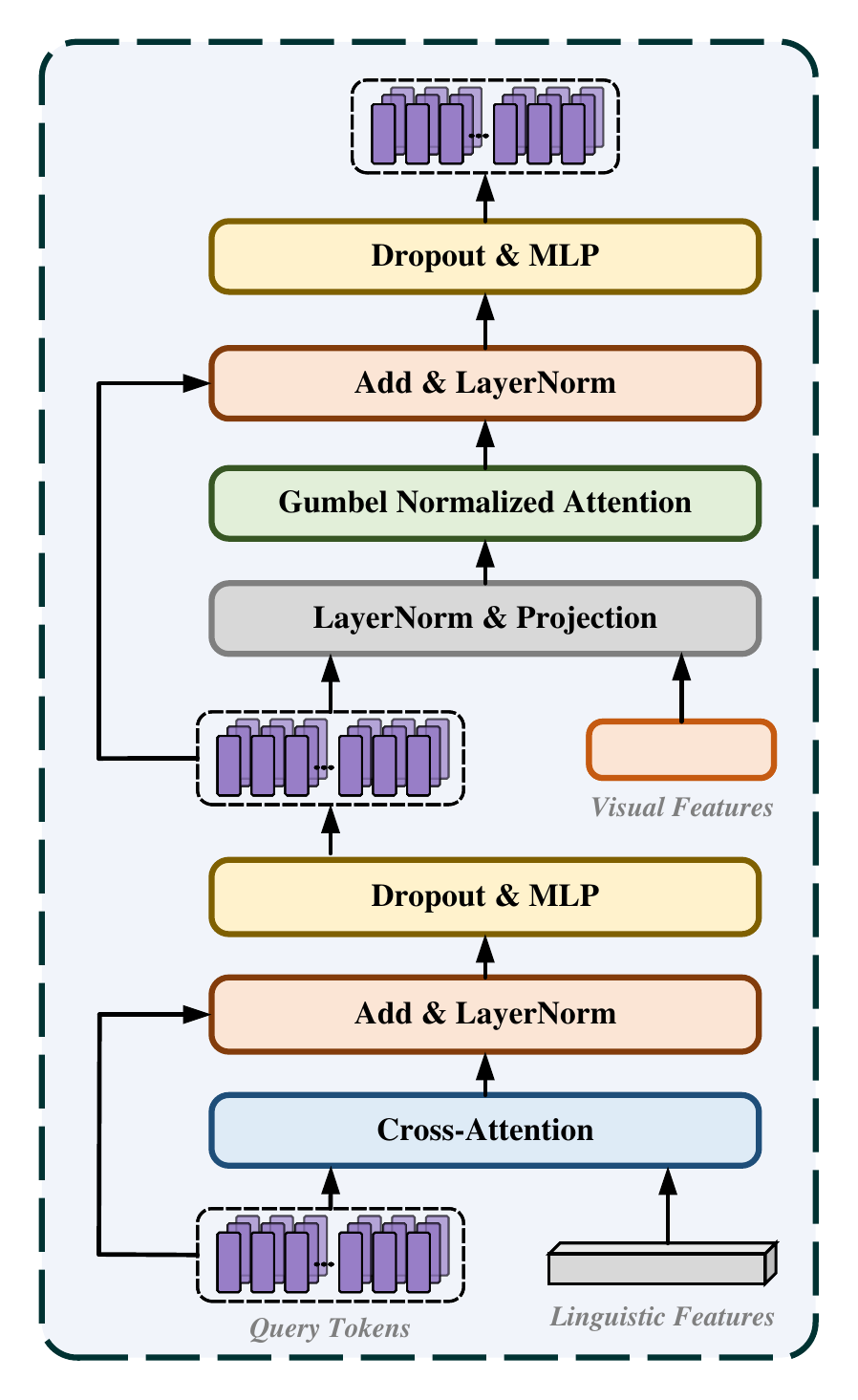}}
		\caption{Structure of the OAQIL within the PCMRD. The OAQIL facilitates cross-modal reasoning through a series of attention mechanisms: cross-attention between query tokens and linguistic features, followed by Gumbel normalized attention for query-vision feature matching. Multi-layer perceptron (MLP) and normalization layers ensure effective feature integration.}\label{OAQIL}
	\end{center}
\end{figure}

As the core component of PCMRD, the OAQILs facilitate precise visual region localization through query text injection and adaptive query-vision feature matching, as illustrated in Fig.~\ref{OAQIL}. For simplicity, we take the first decoding process as an example. Inspired by the success of query mechanisms in CV field, we introduce learnable query tokens $Q \in \mathbb{R}^{M \times C_q}$ to encode object-centric representations, where $M$ denotes the number of tokens and $C_q$ represents the token dimension. Next, the tokens $Q$ are processed by OAQIL to capture the object-level information and update token features. 

The input query tokens are first aligned with linguistic features through cross-attention, enabling each query token to focus on the relevant linguistic information required at the current layer. The output linguistic-enhanced tokens serve as a foundation for subsequent cross-modal reasoning and interaction:

\begin{equation}
	Q_q=\mathbf{W}_q  Q,  L_k=\mathbf{W}_k  L, L_v=\mathbf{W}_v  L,
\end{equation}
\begin{equation}
	Q_l=\left(\operatorname{softmax}\left(\frac{Q_q\left(L_k\right)^{\top}}{\sqrt{C_q}}\right) L_v\right) \mathbf{W}_c,
\end{equation}where $Q_q$, $L_k$ and $L_v$ represent the query, key, and value in the cross-attention mechanism, respectively, while $\mathbf{W}_c, \mathbf{W}_q \in \mathbb{R}^{C_q \times C_q}$ and $\mathbf{W}_k, \mathbf{W}_v \in \mathbb{R}^{D_l \times C_q}$ are learnable projection matrices. The generated linguistic-enhanced tokens $Q_l$ are subsequently fed into the Gumbel normalized attention mechanism to query and associate relevant visual features with the tokens.

Subsequently, the fused cross-scale visual features $V^{\operatorname{fused}}_{3,4} \in \mathbb{R}^{H_3 \times W_3 \times C^f_1}$and linguistic-enhanced query tokens $Q_l$ undergo cross-modal attention interaction facilitated by the designed Gumbel normalized attention mechanism.

$Q_l$ and $V^{\operatorname{fused}}_{3,4}$ are first projected into a common feature space, and we calculate the similarities $\operatorname{S}_{\text {pixel }} \in \mathbb{R}^{M \times H_3 W_3}$ between them:

\begin{equation}
	\operatorname{S}_{\text {pixel }}=\operatorname{norm}\left(\mathbf{W}_t  Q_l\right)  \operatorname{norm}\left(\operatorname{flatten}\left(\mathbf{W}_d  V^{\operatorname{fused}}_{3,4}\right) \right)^{\top},
\end{equation}where $\mathbf{W}_t \in \mathbb{R}^{C_q \times C_q}$ and $\mathbf{W}_d \in \mathbb{R}^{C^f_1 \times C_q}$ denote the learnable projection matrices. The $\operatorname{flatten}$ operation reshapes the feature $V^{\operatorname{fused}}_{3,4}$ into the visual feature with $H_3W_3$ vectors, while $\operatorname{norm}$ refers to $L_2$ normalization.

Based on the similarities and the learnable Gumbel-softmax \cite{jang2016categorical, maddison2016concrete}, the features in $V^{\operatorname{fused}}_{3,4}$  are then hard-assigned to the tokens $Q_l$, resulting in the generation of updated tokens $Q_u$ through the designed OAQIL:

\begin{equation}
	S_{\text {gumbel }}=\operatorname{softmax}\left(\left(S_{\text {pixel }}+G\right) / \tau\right),
\end{equation}
\begin{equation}
	S_{\text {onehot }}=\operatorname{onehot}\left(\operatorname{argmax}_N\left(S_{\text {gumbel }}\right)\right),
\end{equation}
\begin{equation}
	S_{\text {mask }}=\left(S_{\text {onehot }}\right)^{\top}-\operatorname{sg}\left(S_{\text {gumbel }}\right)+S_{\text {gumbel }},
\end{equation}
\begin{equation}
	Q_u=\operatorname{MLP}(S_{\text {mask}} \cdot \operatorname{flatten}(\mathbf{W}_d V^{\operatorname{fused}}_{3,4}))+ \mathbf{W}_t Q_l
\end{equation}where $G \in \mathbb{R}^{M \times H_3 W_3}$ is sampled from the Gumbel (0,1) distribution, $\tau$ is a learnable coefficient for adjusting the assignment boundary, and $\operatorname{sg}$ denotes the stop gradient operator. The $\operatorname{argmax}_N$ selects the token with the highest similarity for each visual feature, while the one-hot operation encodes these token indices into one-hot vectors $S_{\text{onehot}} \in \mathbb{R}^{H_3 W_3 \times M}$. The resulting mask $S_{\text {mask }} \in \mathbb{R}^{M \times H_3 W_3}$ represents the assignment of features from $V^{\operatorname{fused}}_{3,4}$ to $Q_l$.

\section{Experiments}
\label{section:EXPERIMENTS}

\subsection{Experimental Settings}

\subsubsection{Datasets}
To evaluate the effectiveness of our algorithm, we conducted experiments on three public datasets: the RRSIS-D dataset\cite{liu2024rotated}, the RefSegRS dataset\cite{yuan2024rrsis}, and the RISBench dataset\cite{dong2024cross}. 

The RRSIS-D dataset features a diverse collection of 17,402 images, each paired with referring expressions and corresponding masks. It includes 20 semantic categories enriched by 7 descriptive attributes, enabling detailed semantic context representation. The dataset is organized into training (12,181 triplets), validation (1,740 triplets), and testing subsets (3,481 triplets). Images are standardized at a resolution of 800×800 pixels, emphasizing large-scale and complex spatial scenes.

The RefSegRS dataset comprises 4,420 image-language-label triplets across 285 distinct scenes. The data is divided into training (151 scenes, 2,172 expressions), validation (31 scenes, 431 expressions), and testing subsets (103 scenes, 1,817 expressions). All images are formatted to 512×512 pixels with a spatial resolution of 0.13 meters, making it suitable for fine-grained RRSIS tasks.

The RISBench dataset comprises 52,472 triplets, significantly expanding the scale and diversity of referring image segmentation benchmarks. The dataset is split into training (26,300 triplets), validation (10,013 triplets), and testing (16,159 triplets) subsets. Images, uniformly sized at 512×512 pixels, span spatial resolutions from 0.1 to 30 meters, catering to multi-scale segmentation challenges. Semantic labels are distributed across 26 unique classes, each annotated with 8 descriptive attributes, offering comprehensive semantic and contextual richness.

\subsubsection{Evaluation Metrics}

To evaluate the performance of our algorithm as well as other comparison methods, we employ several standard evaluation metrics. These include the overall Intersection-over-Union (oIoU), mean Intersection-over-Union (mIoU), and precision at different threshold values X $\in$ \{0.5,0.6,0.7,0.8,0.9\} (Pr@X). 

oIoU is calculated as the ratio of the total intersection area to the total union area across all test samples. This metric emphasizes the segmentation performance on larger objects by aggregating the intersection and union areas before computing the ratio:

\begin{equation} 
	\text{oIoU} = \frac{\sum_{t} I_t}{\sum_{t} U_t}
\end{equation}

mIoU, on the other hand, is the average of the IoU scores computed individually for each test sample. This approach ensures that both small and large objects contribute equally to the final metric, providing a balanced evaluation of the model's performance across varying object sizes:

\begin{equation} 
	\text{mIoU} = \frac{1}{K} \sum_{t=1}^{K} \frac{I_t}{U_t} 
\end{equation}where $t$ indexes each image-language-label triplet in the dataset, $K$ denotes the total number of samples, and $I_t$ and $U_t$represent the intersection and union areas between the predicted segmentation mask and the corresponding ground truth annotation for the $t$-th sample.

Pr@X measures the proportion of correctly segmented objects that achieve an IoU greater than or equal to the threshold X. This metric provides insight into the model's ability to produce high-quality segmentations under varying strictness levels.

\begin{table*}[htbp]
	\centering
	\caption{Comparison with state-of-the-art methods on the RRSIS-D dataset. Optimal and sub-optimal performance in each metric are marked by \textcolor{red}{\textbf{red}} and \textcolor{blue}{\textbf{blue}}.}
	\label{rrsisd_comparison}
	\renewcommand\arraystretch{1.4}
	\fontsize{8}{12}\selectfont
	\resizebox{\textwidth}{!}{
		\begin{tabular}{l|c|c|c|c|c|c|c|c|c|c|c|c|c|c|c|c}
			\cmidrule{1-17} 
			\multirow{2}{*}{Method} & \multirow{2}{*}{Visual Encoder} & \multirow{2}{*}{Text Encoder} & \multicolumn{2}{c|}{Pr@0.5} & \multicolumn{2}{c|}{Pr@0.6} & \multicolumn{2}{c|}{Pr@0.7} & \multicolumn{2}{c|}{Pr@0.8} & \multicolumn{2}{c|}{Pr@0.9} & \multicolumn{2}{c|}{oIoU} & \multicolumn{2}{c}{mIoU} \\ \cline{4-17}  % Adds a horizontal line above the Val and Test columns
			& & & Val & Test & Val & Test & Val & Test & Val & Test & Val & Test & Val & Test & Val & Test \\ 
			\cmidrule{1-17} 
			RRN\cite{li2018referring} & ResNet-101 & LSTM & 51.09 & 51.07 & 42.47 & 42.11 & 33.04 & 32.77 & 20.80 & 21.57 & 6.14 & 6.37 & 66.53 & 66.43 & 46.06 & 45.64 \\
			CSMA\cite{ye2019cross} & ResNet-101 & None & 55.68 & 55.32 & 48.04 & 46.45 & 38.27 & 37.43 & 26.55 & 25.39 & 9.02 & 8.15 & 69.68 & 69.43 & 48.85 & 48.54 \\
			LSCM\cite{hui2020linguistic} & ResNet-101 & LSTM & 57.12 & 56.02 & 48.04 & 46.25 & 37.87 & 37.70 & 26.35 & 25.28 & 7.93 & 7.86 & 69.28 & 69.10 & 50.36 & 49.92 \\
			CMPC\cite{huang2020referring} & ResNet-101 & LSTM & 57.93 & 55.83 & 48.85 & 47.40 & 36.94 & 35.28 & 25.25 & 25.45 & 9.31 & 9.20 & 70.15 & 69.41 & 51.01 & 49.24 \\
			BRINet\cite{hu2020bi} & ResNet-101 & LSTM & 58.79 & 56.90 & 49.54 & 48.77 & 39.65 & 38.61 & 28.21 & 27.03 & 9.19 & 8.93 & 70.73 & 69.68 & 51.41 & 49.45 \\
			CMPC+\cite{liu2021cross} & ResNet-101 & LSTM & 59.19 & 57.95 & 49.41 & 48.31 & 38.67 & 37.61 & 25.91 & 24.33 & 8.16 & 7.94 & 70.80 & 70.13 & 51.63 & 50.12 \\ \hline	
			
			CRIS\cite{wang2022cris} & ResNet-101 & CLIP & 56.44	& 54.84	& 47.87	& 46.77	& 39.77	& 38.06	& 29.31	& 28.15	& 11.84	& 11.52	& 70.98	& 70.46	& 50.75 & 49.69 \\
			BKINet\cite{ding2023bilateral} & ResNet-101 & CLIP & 58.79 &	56.9 & 49.54 & 48.77 & 39.65 & 39.12 & 28.21 & 27.03 & 9.19 & 9.16 & 70.78 & 69.89 & 51.14 & 49.65 \\
			ETRIS\cite{xu2023bridging} & ResNet-101 & CLIP & 62.10	& 61.07	& 53.73	& 50.99	& 43.12	& 40.94	& 30.79	& 29.30 & 12.90 & 11.43 & 72.75 & 71.06 & 55.21 & 54.21 \\ \hline	
			
			SLViT\cite{ouyang2023slvit} & Swin-B & BERT & 41.44 & 37.89 & 33.74 & 31.20 & 26.09 & 24.56 & 17.59 & 16.60 & 6.15 & 6.55 & 61.27 & 59.80 & 38.05 & 35.79 \\
			robust-ref-seg\cite{wu2024towards} & Swin-B & BERT & 64.22 & 66.59 & 58.72 & 59.58 & 50.00 & 49.93 & 35.78 & 38.72 & 24.31 & 23.30 & 76.39 & \textcolor{red}{\textbf{77.40}} & 58.92 & 58.91 \\
			LAVT\cite{yang2022lavt} & Swin-B & BERT & 65.23 & 63.98 & 58.79 & 57.57 & 50.29 & 49.30 & 40.11 & 38.06 & 23.05 & 22.29 & 76.27 & 76.16 & 57.72 & 56.82 \\
			CrossVLT\cite{cho2023cross} & Swin-B & BERT & 67.07 & 66.42 & 59.54 & 59.41 & 50.80 & 49.76 & 40.57 & 38.67 & 23.51 & 23.30 & 76.25 & 75.48 & 59.78 & 58.48 \\
			LGCE\cite{yuan2024rrsis} & Swin-B & BERT & 68.10 & 67.65 & 60.61 & 61.53 & 51.45 & 51.42 & 42.34 & 39.62 & 23.85 & 22.94 & 76.68 & 76.33 & 60.16 & 59.37 \\
			RMSIN\cite{liu2024rotated} & Swin-B & BERT & 68.39 & 67.16 & 61.72 & 60.36 & 52.24 & 50.16 & 41.44 & 38.72 & 23.16 & 22.81 & \textcolor{red}{\textbf{77.53}} & 75.79 & 60.23& 58.79 \\
			RIS-DMMI\cite{hu2023beyond} & Swin-B & BERT & 70.40 & 68.74	& 63.05	& 60.96	& 54.14	& 50.33	& 41.95	& 38.38	& 23.85 & 21.63	& 77.01	& 76.20 & 61.70	& 60.25 \\
			
			CARIS\cite{liu2023caris} & Swin-B & BERT & \textcolor{blue}{\textbf{71.61}} & \textcolor{blue}{\textbf{71.50}} & \textcolor{blue}{\textbf{64.66}} & \textcolor{blue}{\textbf{63.52}} & 54.14 & \textcolor{blue}{\textbf{52.92}} & 42.76 & \textcolor{blue}{\textbf{40.94}} & 23.79 & 23.90 & \textcolor{blue}{\textbf{77.48}} & \textcolor{blue}{\textbf{77.17}} & \textcolor{blue}{\textbf{62.88}} & \textcolor{blue}{\textbf{62.12}} \\ 	
			%			CroBIM\cite{dong2024cross} & Swin-B & BERT & \textcolor{blue}{\textbf{74.20}} & \textcolor{red}{\textbf{75.00}} & \textcolor{blue}{\textbf{66.15}} & \textcolor{blue}{\textbf{66.32}} & 54.08 & \textcolor{blue}{\textbf{54.31}} & 41.38 & \textcolor{blue}{\textbf{41.09}} & 22.30 & 21.78 & 76.24 & 76.37 & \textcolor{blue}{\textbf{63.99}} & \textcolor{blue}{\textbf{64.24}}\\  
			\hline
			
			EVP\cite{lavreniuk2023evp} & Stable Diffusion & CLIP & 57.76 & 56.42 & 51.49 & 50.01 & 42.87 & 42.26 & 33.97 & 33.81 & 19.14 & 19.16 & 73.54 & 73.08 & 51.77 & 50.85  \\
			VPD\cite{zhao2023unleashing} & Stable Diffusion & CLIP & 70.92 & 68.43 & 63.74 & 62.17 & \textcolor{blue}{\textbf{54.94}} & 52.26 & \textcolor{blue}{\textbf{44.37}} & 40.85 & \textcolor{blue}{\textbf{25.23}} & \textcolor{blue}{\textbf{25.31}} & 77.04 & 75.85 & 61.91 & 60.31 \\
			DiffRIS (ours)  & Stable Diffusion & CLIP & \textcolor{red}{\textbf{73.45}} & \textcolor{red}{\textbf{71.79}} & \textcolor{red}{\textbf{66.55}} & \textcolor{red}{\textbf{65.47}} & \textcolor{red}{\textbf{56.72}} & \textcolor{red}{\textbf{54.64}} & \textcolor{red}{\textbf{44.71}} & \textcolor{red}{\textbf{42.89}} & \textcolor{red}{\textbf{27.01}} & \textcolor{red}{\textbf{26.46}} & 76.84 & 76.05 & \textcolor{red}{\textbf{63.64}} & \textcolor{red}{\textbf{62.19}} \\
			\hline
		\end{tabular}
	}
\end{table*}

\subsubsection{Training Details}

The model was implemented and trained using the PyTorch framework, leveraging a hardware setup of eight NVIDIA A800 GPUs, each equipped with 80GB of memory. Besides, the model was optimized using the AdamW optimizer with a learning rate of $3\times10^{-5}$ , a weight decay of 0.01, and a batch size of 32. The training process is conducted over 40 epochs to ensure sufficient optimization and convergence. During both training and evaluation, input images were resized to a resolution of 512$\times$512 pixels. Notably, no data augmentation or post-processing techniques were employed throughout the entire pipeline.

\begin{table*}[htbp]
	\centering
	\caption{Comparison with state-of-the-art methods on the RefSegRS dataset. Optimal and sub-optimal performance in each metric are marked by \textcolor{red}{\textbf{red}} and \textcolor{blue}{\textbf{blue}}.}
	\label{refsegrs_comparison}
	\renewcommand\arraystretch{1.4}
	\fontsize{8}{12}\selectfont
	\resizebox{\textwidth}{!}{
		\begin{tabular}{l|c|c|c|c|c|c|c|c|c|c|c|c|c|c|c|c}
			\cmidrule{1-17} 
			\multirow{2}{*}{Method} & \multirow{2}{*}{Visual Encoder} & \multirow{2}{*}{Text Encoder} & \multicolumn{2}{c|}{Pr@0.5} & \multicolumn{2}{c|}{Pr@0.6} & \multicolumn{2}{c|}{Pr@0.7} & \multicolumn{2}{c|}{Pr@0.8} & \multicolumn{2}{c|}{Pr@0.9} & \multicolumn{2}{c|}{oIoU} & \multicolumn{2}{c}{mIoU} \\ \cline{4-17}  % Adds a horizontal line above the Val and Test columns
			& & & Val & Test & Val & Test & Val & Test & Val & Test & Val & Test & Val & Test & Val & Test \\ 
			\cmidrule{1-17} 
			BRINet\cite{hu2020bi} & ResNet-101 & LSTM & 36.86 & 20.72 & 35.53 & 14.26 & 19.93 & 9.87 & 10.66 & 2.98 & 2.84 & 1.14 & 61.59 & 58.22 & 38.73  & 31.51 \\
			CMSA\cite{ye2019cross} & ResNet-101 & None & 39.24 & 26.14 & 38.44 & 18.52 & 20.39  & 10.66 & 11.79 & 4.71 & 1.52 & 0.69 & 63.84 & 62.11 & 43.62 & 38.72 \\
			MAttNet\cite{yu2018mattnet} & ResNet-101 & LSTM & 48.56 & 28.79 & 40.26 & 22.51 & 20.59 & 11.32 & 12.98 & 3.62 & 2.02 & 0.79 & 66.84 & 64.28 & 41.73 & 33.42 \\
			RRN\cite{li2018referring} & ResNet-101 & LSTM & 55.43 & 30.26 & 42.98 & 23.01 & 23.11 & 14.87 & 13.72 & 7.17 & 2.64 & 0.98 & 69.24 & 65.06 &  50.81 & 41.88 \\
			LSCM\cite{hui2020linguistic} & ResNet-101 & LSTM  & 56.82 & 31.54 & 41.24 & 20.41 & 21.85 & 9.51 & 12.11 & 5.29 & 2.51 & 0.84 & 62.82 & 61.27 & 40.59 & 35.54 \\ \hline
			
			BKINet\cite{ding2023bilateral} & ResNet-101 & CLIP & 52.04 & 36.12 & 35.31 & 20.62 & 18.35 & 15.22 & 12.78 & 6.26 & 1.23 & 1.33 & 75.37 & 63.37 & 56.12 & 40.41 \\
			CRIS\cite{wang2022cris} & ResNet-101 & CLIP & 53.13 & 35.77 & 36.19 & 24.11 & 24.36 & 14.36 & 11.83 & 6.38 & 2.55 & 1.21 & 72.14 & 65.87 & 53.74 & 43.26 \\
			ETRIS\cite{xu2023bridging} & ResNet-101 & CLIP & 54.99 & 35.77 & 35.03 & 23.00 & 25.06 & 13.98 & 12.53 & 6.44 & 1.62 & 1.10 & 72.89 & 65.96 & 54.03 & 43.11 \\ \hline
			
			CrossVLT\cite{cho2023cross} & Swin-B & BERT & 67.52	& 41.94 &	43.85 & 25.43 & 25.99 & 15.19 & 14.62 & 3.71 & 1.87 & 1.76 & 76.12 & 69.73 & 55.27 & 42.81 \\
			RMSIN\cite{liu2024rotated} & Swin-B & BERT & 68.21 & 42.32 & 46.64 & 25.87 & 24.13 & 14.20 & 13.69 & 6.77 & 3.25 & 1.27 & 74.40 & 68.31 & 54.24 & 42.63 \\
			CARIS\cite{liu2023caris} & Swin-B & BERT & 68.45 & 45.40 & 47.10 & 27.19 & 25.52 & 15.08 & 14.62 & 7.87 & 3.71 & 1.98 & 75.79 & 69.74 & 54.30 & 42.66 \\
			LGCE\cite{yuan2024rrsis} & Swin-B & BERT & 79.81	& 50.19 & 54.29	& 28.62 & 29.70 & 17.17 & 15.31 & 9.36 & 5.10 & 2.15 & 78.24 & 71.59 & 60.66 & 46.57 \\ 
			LAVT\cite{yang2022lavt} & Swin-B & BERT & 80.97	& 51.84 & 58.70 & 30.27 & 31.09 & 17.34 & 15.55 & 9.52 & 4.64 & 2.09 & 78.50 & 71.86 & 61.53 & 47.40
			\\ 
			robust-ref-seg\cite{wu2024towards} & Swin-B & BERT & 81.67 & 50.25 & 52.44 & 28.01 & 30.86 & 17.83 & 17.17 & 9.19 & 5.80 & 2.48 & 77.74 & 71.13 & 60.44 & 47.12 \\
			RIS-DMMI\cite{hu2023beyond} & Swin-B & BERT & 86.17 & 63.89 & 74.71 & 44.30 & 38.05 & 19.81 & 18.10 & 6.49 & 3.25 & 1.00 & 74.02 & 68.58 & 65.72 & 52.15 \\
			CroBIM\cite{dong2024cross} & Swin-B & BERT & \textcolor{blue}{\textbf{87.24}} & 64.83 & 75.17 & 44.41 & 44.78 & 17.28 & 19.03 & 9.69 & 6.26 & 2.20 &  78.85 & 72.30 & 65.79 & 52.69 \\	
			
			\hline
			
			EVP\cite{lavreniuk2023evp} & Stable Diffusion & CLIP & 57.77 & 32.42 & 38.05 & 22.62 & 28.54 & 16.02 & 17.87 & 9.30 & 6.50 & 2.04 & 77.14 & 69.56 & 53.95 & 38.07 \\
			VPD\cite{zhao2023unleashing} & Stable Diffusion & CLIP & 86.77 & \textcolor{blue}{\textbf{69.24}} & \textcolor{blue}{\textbf{76.57}} & \textcolor{blue}{\textbf{51.29}} & \textcolor{blue}{\textbf{46.40}} & \textcolor{blue}{\textbf{33.74}} & \textcolor{blue}{\textbf{30.39}} & \textcolor{blue}{\textbf{19.59}} & \textcolor{blue}{\textbf{12.99}} & \textcolor{red}{\textbf{6.33}} & \textcolor{blue}{\textbf{83.38}} & \textcolor{red}{\textbf{76.24}} & \textcolor{blue}{\textbf{68.00}} & \textcolor{blue}{\textbf{57.30}}  \\
			DiffRIS (ours)  & Stable Diffusion & CLIP & \textcolor{red}{\textbf{94.66}} & \textcolor{red}{\textbf{74.57}} & \textcolor{red}{\textbf{92.11}} & \textcolor{red}{\textbf{64.56}} & \textcolor{red}{\textbf{84.22}} & \textcolor{red}{\textbf{46.51}} & \textcolor{red}{\textbf{57.08}} & \textcolor{red}{\textbf{24.49}} & \textcolor{red}{\textbf{14.62}} & \textcolor{blue}{\textbf{5.78}} & \textcolor{red}{\textbf{85.63}} & \textcolor{blue}{\textbf{76.18}} & \textcolor{red}{\textbf{78.28}} & \textcolor{red}{\textbf{62.38}}\\ \hline
			
		\end{tabular}
	}
\end{table*}

\begin{table*}[htbp]
	\centering
	\caption{Comparison with state-of-the-art methods on the RISBench dataset. Optimal and sub-optimal performance in each metric are marked by \textcolor{red}{\textbf{red}} and \textcolor{blue}{\textbf{blue}}.}
	\label{risbench_comparison}
	\renewcommand\arraystretch{1.4}
	\fontsize{8}{12}\selectfont
	\resizebox{\textwidth}{!}{
		\begin{tabular}{l|c|c|c|c|c|c|c|c|c|c|c|c|c|c|c|c}
			\cmidrule{1-17} 
			\multirow{2}{*}{Method} & \multirow{2}{*}{Visual Encoder} & \multirow{2}{*}{Text Encoder} & \multicolumn{2}{c|}{Pr@0.5} & \multicolumn{2}{c|}{Pr@0.6} & \multicolumn{2}{c|}{Pr@0.7} & \multicolumn{2}{c|}{Pr@0.8} & \multicolumn{2}{c|}{Pr@0.9} & \multicolumn{2}{c|}{oIoU} & \multicolumn{2}{c}{mIoU} \\ \cline{4-17}  % Adds a horizontal line above the Val and Test columns
			& & & Val & Test & Val & Test & Val & Test & Val & Test & Val & Test & Val & Test & Val & Test \\ 
			\cmidrule{1-17} 
			BRINet\cite{hu2020bi} & ResNet-101 & LSTM & 52.11 & 52.87 & 45.17 & 45.39 & 37.98 & 38.64 & 30.88 & 30.79 & 10.28 & 11.86 & 46.27 & 48.73 & 41.54 & 42.91 \\
			RRN\cite{li2018referring} & ResNet-101 & LSTM & 54.62 & 55.04 & 46.88 & 47.31 & 39.57 & 39.86 & 32.64 & 32.58 & 11.57 & 13.24 & 47.28 & 49.67 & 42.65 & 43.18 \\
			CMPC\cite{huang2020referring} & ResNet-101 & LSTM & 54.89 & 55.17 & 47.77 & 47.84 & 40.38 & 40.28 & 32.89 & 32.87 & 12.63 & 14.55 & 47.59 & 50.24 & 42.83 & 43.82\\
			LSCM\cite{hui2020linguistic} & ResNet-101 & LSTM  & 55.87 & 55.26 & 47.24 & 47.14 & 40.22 & 40.10 & 33.55 & 33.29 & 12.78 & 13.91 & 47.99 & 50.08 & 43.21 & 43.69 \\
			
			MAttNet\cite{yu2018mattnet} & ResNet-101 & LSTM & 56.77 & 56.83 & 48.51 & 48.02 & 41.53 & 41.75 & 34.33 & 34.18 & 13.84 & 15.26 & 48.66 & 51.24 & 44.28 & 45.71 \\
			
			CMPC+\cite{liu2021cross} & ResNet-101 & LSTM & 57.84 & 58.02 & 49.24 & 49.00 & 42.34 & 42.53 & 35.77 & 35.26 & 14.55 & 17.88 & 50.29 & 53.98 & 45.81 & 46.73 \\	
			\hline
			
			ETRIS\cite{xu2023bridging} & ResNet-101 & CLIP & 59.87 & 60.98 & 49.91 & 51.88 & 35.88 & 39.87 & 20.10 & 24.49 & 8.54 & 11.18 & 64.09 & 67.61 & 51.13 & 53.06 \\
			CRIS\cite{wang2022cris} & ResNet-101 & CLIP & 63.42 & 63.67 & 54.32 & 55.73 & 41.15 & 44.42 & 24.66 & 28.80 & 10.27 & 13.27 & 66.26 & 69.11 & 53.64 & 55.18 \\
			\hline
			
			robust-ref-seg\cite{wu2024towards} & Swin-B & BERT & 67.42 & 69.15 & 61.72 & 63.24 & 53.64 & 55.33 & 40.71 & 43.27 & 19.43 & 24.20 & 69.50 & 74.23 & 59.37 & 61.25 \\
			LAVT\cite{yang2022lavt} & Swin-B & BERT & 68.27 & 69.40 & 62.71 & 63.66 & 54.46 & 56.10 & 43.13 & 44.95 & 21.61 & 25.21 & 69.39 & 74.15 & 60.45 & 61.93 \\
			LGCE\cite{yuan2024rrsis} & Swin-B & BERT & 68.20 & 69.64 & 62.91 & 64.07 & 55.01 & 56.26 & 43.38 & 44.92 & 21.58 & 25.74 & 68.81 & 73.87 & 60.44 & 62.13 \\ 
			RMSIN\cite{liu2024rotated} & Swin-B & BERT & 70.05 & 71.01 & 64.64 & 65.46 & 56.37 & 57.69 & 44.14 & 45.50 & 21.40 & 25.92 & 69.51 & 74.09 & 61.78 & 63.07 \\
			CrossVLT\cite{cho2023cross} & Swin-B & BERT & 70.05 & 70.62 & 64.29 & 65.05 & 56.97 & 57.40 & 44.49 & 45.80 & 21.47 & 26.10 & 69.77 & 74.33 & 61.54 & 62.84 \\ 
			RIS-DMMI\cite{hu2023beyond} & Swin-B & BERT & 71.27 & 72.05 & 66.02 & 66.48 & 58.22 & 59.07 & 45.57 & 47.16 & 22.43 & 26.57 & \textcolor{red}{\textbf{70.58}} & \textcolor{blue}{\textbf{74.82}} & 62.62 & 63.93 \\
			CARIS\cite{liu2023caris} & Swin-B & BERT & \textcolor{blue}{\textbf{73.46}} & \textcolor{blue}{\textbf{73.94}} & \textcolor{blue}{\textbf{68.51}} & \textcolor{blue}{\textbf{68.93}} & \textcolor{blue}{\textbf{60.92}} & \textcolor{blue}{\textbf{62.08}} & \textcolor{blue}{\textbf{48.47}} & \textcolor{blue}{\textbf{50.31}} & \textcolor{blue}{\textbf{24.98}} & \textcolor{blue}{\textbf{29.08}} & \textcolor{blue}{\textbf{70.55}} & \textcolor{red}{\textbf{75.10}} & \textcolor{blue}{\textbf{64.40}} & \textcolor{blue}{\textbf{65.79}} \\
			%			CroBIM\cite{dong2024cross} & Swin-B & BERT & \textcolor{blue}{\textbf{76.59}} & \textcolor{blue}{\textbf{75.75}} & \textcolor{blue}{\textbf{71.73}} & \textcolor{blue}{\textbf{70.34}} & \textcolor{blue}{\textbf{64.32}} & \textcolor{blue}{\textbf{63.12}} & \textcolor{blue}{\textbf{53.18}} & \textcolor{blue}{\textbf{51.12}} & \textcolor{blue}{\textbf{28.53}} & 28.45 & 69.08 & 73.61 & \textcolor{blue}{\textbf{67.52}} & \textcolor{blue}{\textbf{67.32}} \\	
			
			\hline
			
			EVP\cite{lavreniuk2023evp} & Stable Diffusion & CLIP & 51.89 & 53.71 & 45.86 & 47.43 & 37.83 & 40.00 & 27.11 & 29.74 & 12.36 & 16.11 & 64.64 & 67.72 & 46.82 & 49.19 \\
			VPD\cite{zhao2023unleashing} & Stable Diffusion & CLIP &  70.30 & 71.51 & 65.08 & 66.19 & 58.20 & 59.29 & 46.13 & 48.21 & 24.59 & 28.88 & 69.62 & 74.47 & 62.25 & 63.81 \\
			DiffRIS (ours)  & Stable Diffusion & CLIP & \textcolor{red}{\textbf{73.99}} & \textcolor{red}{\textbf{74.50}} & \textcolor{red}{\textbf{69.36}} & \textcolor{red}{\textbf{69.42}} & \textcolor{red}{\textbf{63.26}} & \textcolor{red}{\textbf{62.96}} & \textcolor{red}{\textbf{52.77}} & \textcolor{red}{\textbf{53.26}} & \textcolor{red}{\textbf{31.81}} & \textcolor{red}{\textbf{34.57}} & 68.80 & 73.00 & \textcolor{red}{\textbf{66.32}} & \textcolor{red}{\textbf{67.04}}\\ \hline
		\end{tabular}
	}
\end{table*}

\subsection{Performance Analysis}

To evaluate the effectiveness of the proposed approach, we conducted comparative experiments using a selection of state-of-the-art frameworks, including ResNet-LSTM\cite{li2018referring,ye2019cross,hui2020linguistic,huang2020referring,hu2020bi,liu2021cross}, ResNet-CLIP\cite{wang2022cris,ding2023bilateral,xu2023bridging}, Swin-Bert\cite{ouyang2023slvit,yuan2024rrsis,yang2022lavt,liu2024rotated,cho2023cross,hu2023beyond,wu2024towards,liu2023caris}, and Stable Diffusion-CLIP\cite{lavreniuk2023evp,zhao2023unleashing}, as benchmarks.

\subsubsection{Quantitative Results}

Tables~\ref{rrsisd_comparison}--\ref{risbench_comparison} present a comprehensive comparison of RIS performance across the RRSIS-D, RefSegRS, and RisBench datasets. Specifically, the tables report precision at various threshold values (Pr@X), as well as oIoU and mIoU metrics, for both the validation and test sets. The optimal and sub-optimal performances are distinctly highlighted in \textcolor{red}{\textbf{red}} and \textcolor{blue}{\textbf{blue}}, respectively.

\begin{figure*}[tbp]
	\begin{center}
		\centerline{\includegraphics[width=0.95\linewidth]{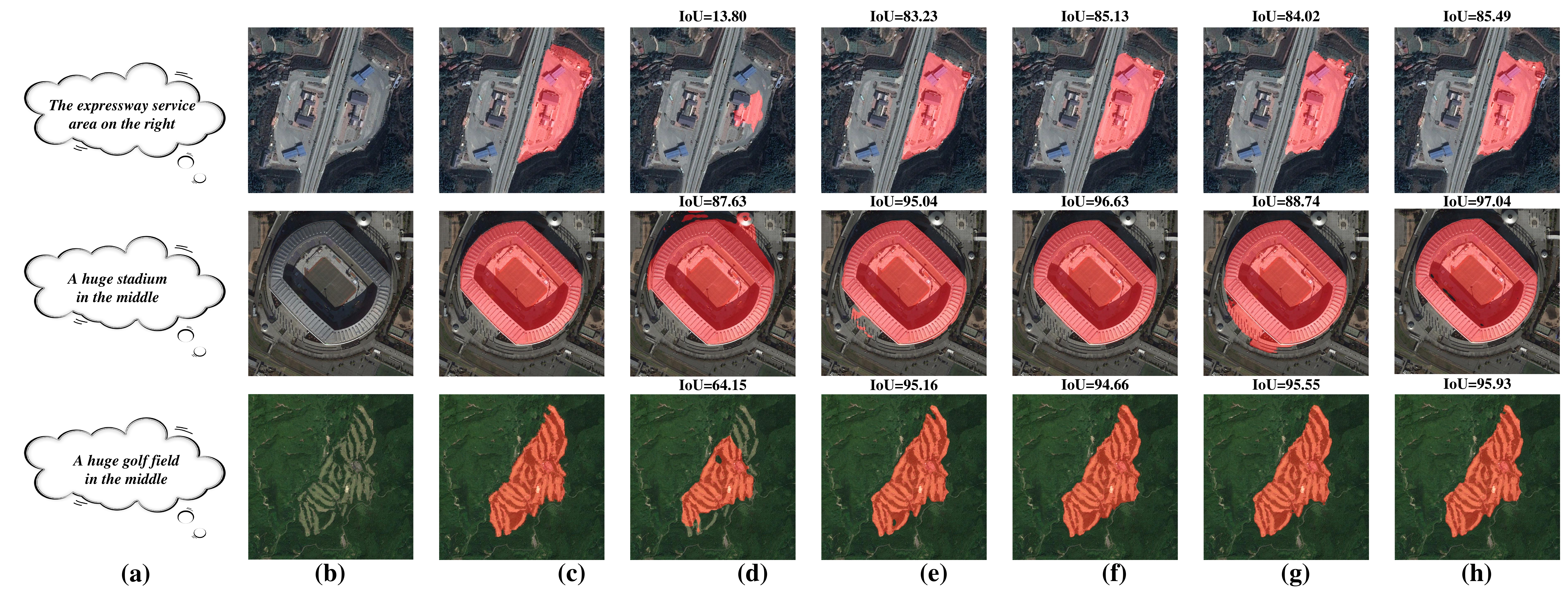}}
		
		\caption{Visualization of segmentation results for DiffRIS and comparison methods on the RRSIS-D dataset test set, with corresponding IoU scores displayed. (a) Query Expressions; (b) Images; (c) Ground Truths; (d) EVP; (e) LAVT; (f) RMSIN; (g) CARIS; (h) DiffRIS (ours)}\label{vis_rrsisd}
	\end{center}
\vspace{-1.1cm}
\end{figure*}
\begin{figure*}[tbp]
	\begin{center}
		\centerline{\includegraphics[width=0.95\linewidth]{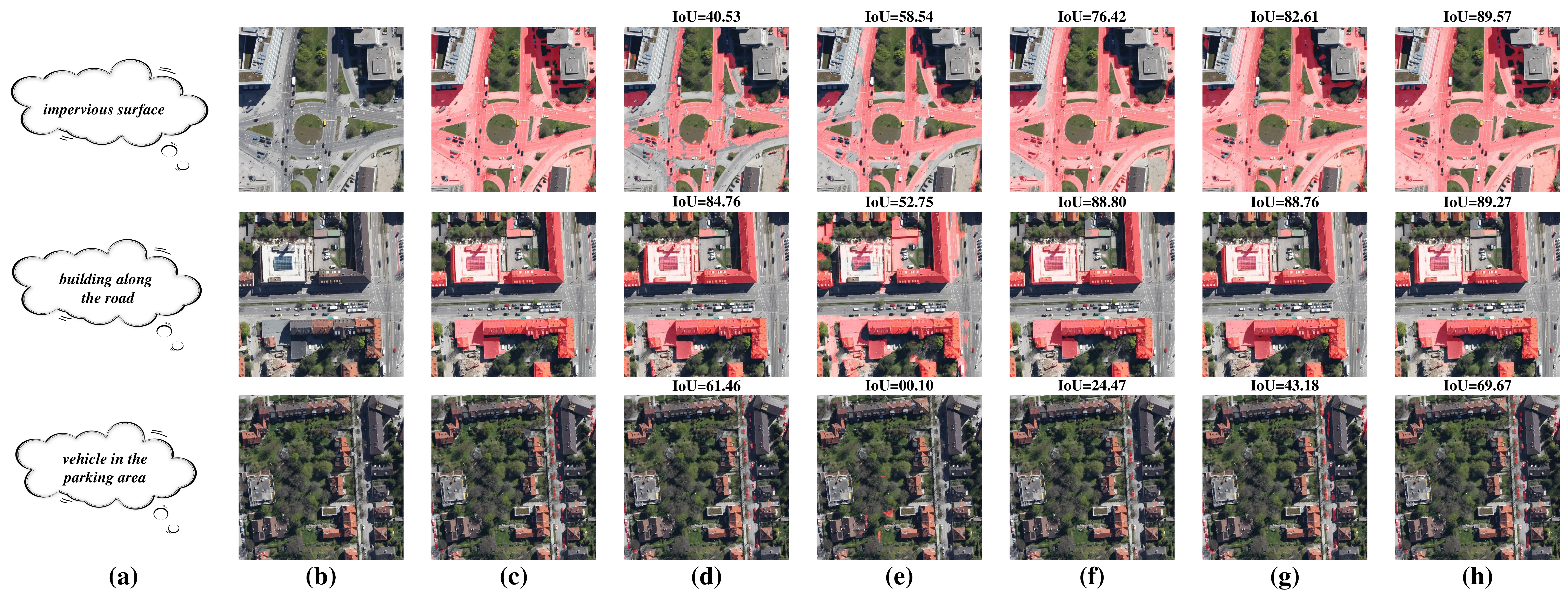}}
		
		\caption{Visualization of segmentation results for DiffRIS and comparison methods on the RefSegRS dataset test set, with corresponding IoU scores displayed. (a) Query Expressions; (b) Images; (c) Ground Truths; (d) LGCE; (e) EVP; (f) VPD; (g) ETRIS; (h) DiffRIS (ours)}\label{vis_refsegrs}
	\end{center}
\vspace{-1.1cm}
\end{figure*}
\begin{figure*}[tbp]
	\begin{center}
		\centerline{\includegraphics[width=0.95\linewidth]{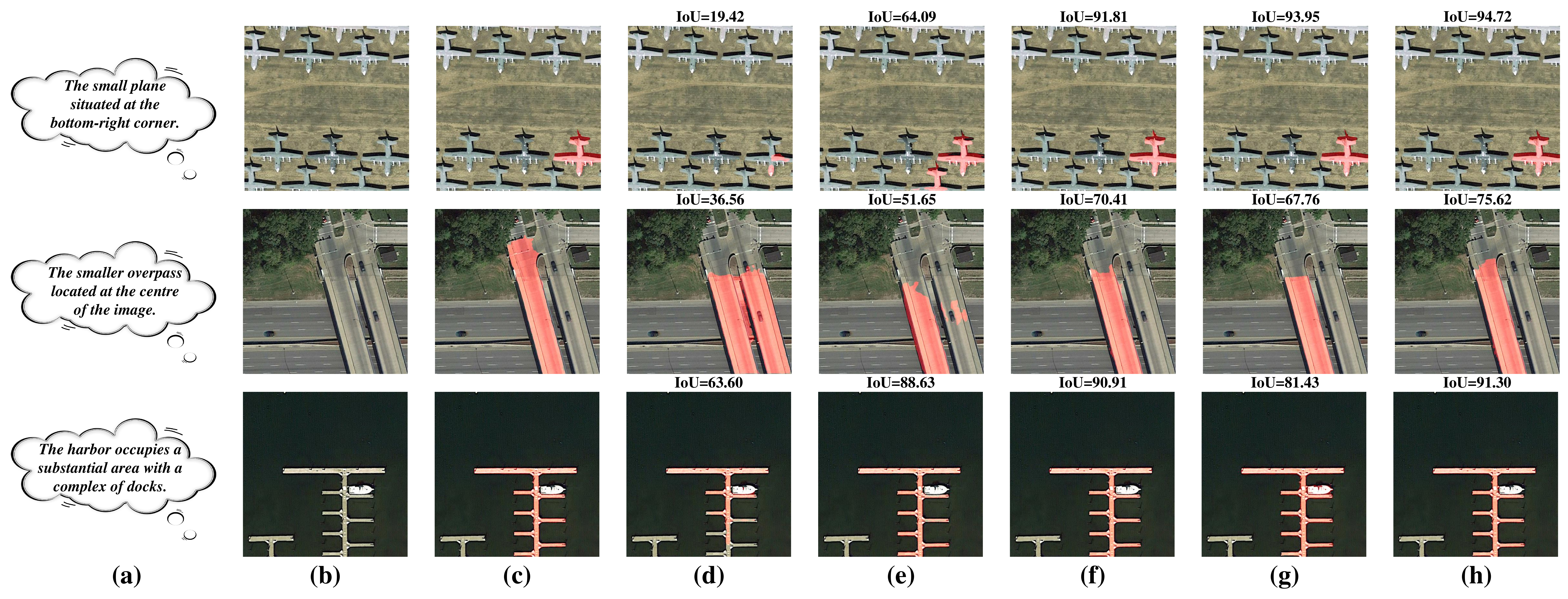}}
		
		\caption{Visualization of segmentation results for DiffRIS and comparison methods on the RisBench dataset test set, with corresponding IoU scores displayed. (a) Query Expressions; (b) Images; (c) Ground Truths; (d) EVP; (e) LAVT; (f) VPD; (g) CrossVLT; (h) DiffRIS (ours)}\label{vis_risbench}
	\end{center}
\end{figure*}

On the RRSIS-D dataset, the proposed DiffRIS framework demonstrates outstanding performance, consistently outperforming all state-of-the-art methods across most metrics. DiffRIS achieves Pr@0.5 values of 73.45 (Val) and 71.79 (Test), surpassing the next best competitor, CARIS, by 1.84\% and 0.29\%, respectively. This advantage extends across higher thresholds (Pr@0.6 to Pr@0.9), underscoring its robustness in maintaining high precision under stricter conditions. Additionally, in terms of mIoU, DiffRIS achieves 63.64 (Val) and 62.19 (Test), further surpassing CARIS by 0.76\% and 0.07\%, respectively. These results underline its ability to handle diverse and complex segmentation tasks effectively.

On the RefSegRS dataset, DiffRIS achieves remarkable superiority, consistently securing optimal results across all metrics. Notably, it achieves Pr@0.5 values of 94.66 (Val) and 74.57 (Test), outperforming the second-best method, VPD, by a substantial margin of 7.89\% and 5.33\%, respectively. This dominance extends to higher thresholds (Pr@0.7 and Pr@0.8), demonstrating the framework’s precision and resilience in challenging scenarios. Furthermore, DiffRIS records the highest mIoU scores of 78.28 (Val) and 62.38 (Test), surpassing VPD by 10.28\% and 5.08\%, respectively. Its oIoU score of 85.63 (Val) is also the highest among all methods, emphasizing its robustness in handling complex spatial contexts.

On the RISBench dataset, DiffRIS continues to exhibit exceptional performance, surpassing all existing methods. It achieves Pr@0.5 values of 73.99 (Val) and 74.50 (Test), outperforming CARIS by 0.53\% and 0.56\%, respectively. This trend is consistently observed at higher thresholds, such as Pr@0.8 and Pr@0.9, indicating DiffRIS’s capability to precisely segment objects under stringent overlap requirements. In terms of segmentation quality, DiffRIS achieves the highest mIoU scores of 66.32 (Val) and 67.04 (Test), surpassing CARIS by 1.92\% and 1.25\%, respectively. The competitive oIoU results further validate its global consistency and segmentation effectiveness.

Across all three datasets, DiffRIS exhibits unparalleled performance, attributed to its innovative integration of Stable Diffusion and CLIP models. This approach effectively enhances cross-modal alignment and enables multi-scale feature representation, resulting in robust, precise, and generalizable segmentation capabilities.

\subsubsection{Qualitative Results}

Figs.\ref{vis_rrsisd}--\ref{vis_risbench} present the visual results of the RRSIS-D, RefSegRS, and RISBench datasets, supplemented by the corresponding IoU scores for each segmentation output. These visual comparisons highlight the superior performance of the proposed DiffRIS framework, which consistently produces more accurate and cohesive segmentation results across diverse and challenging scenarios. Notably, DiffRIS excels in handling intricate cases involving small or overlapping objects, occlusions, and ambiguous boundaries. In contrast to competing methods, DiffRIS demonstrates a remarkable ability to preserve fine-grained details while maintaining global consistency. Furthermore, it effectively mitigates common segmentation artifacts such as under-segmentation and over-segmentation, leading to significantly improved visual coherence. Overall, the visual evidence underscores DiffRIS's ability to accurately model complex spatial and textual relationships, highlighting its efficacy and generalizability across a wide range of datasets and RRSIS applications.

\subsection{Ablation Studies}

In this section, we conduct extensive ablation experiments on the RRSIS-D test set to demonstrate the effectiveness of each component within our DiffRIS.

\subsubsection{Effects on Different Pre-trained Weights}

In this study, we investigate the impact of pre-trained weights from various versions of the Stable Diffusion model on the performance of our DiffRIS framework. Given that DiffRIS is built upon pre-trained text-to-image diffusion models, understanding the influence of different pre-trained weights on model performance is of paramount importance. Specifically, we compare the performance of DiffRIS using weights from four different versions of the Stable Diffusion model: SD-1-1, SD-1-2, SD-1-4, and SD-1-5. We exclude SD-1-3 due to its training iteration count being 30K fewer than SD-1-4. The primary distinction among these versions lies in the number of pre-training iterations conducted on 512×512 resolution images, with higher versions benefiting from more extensive training.

\begin{figure}[tbp]
	\begin{center}
		\centerline{\includegraphics[width=1\linewidth]{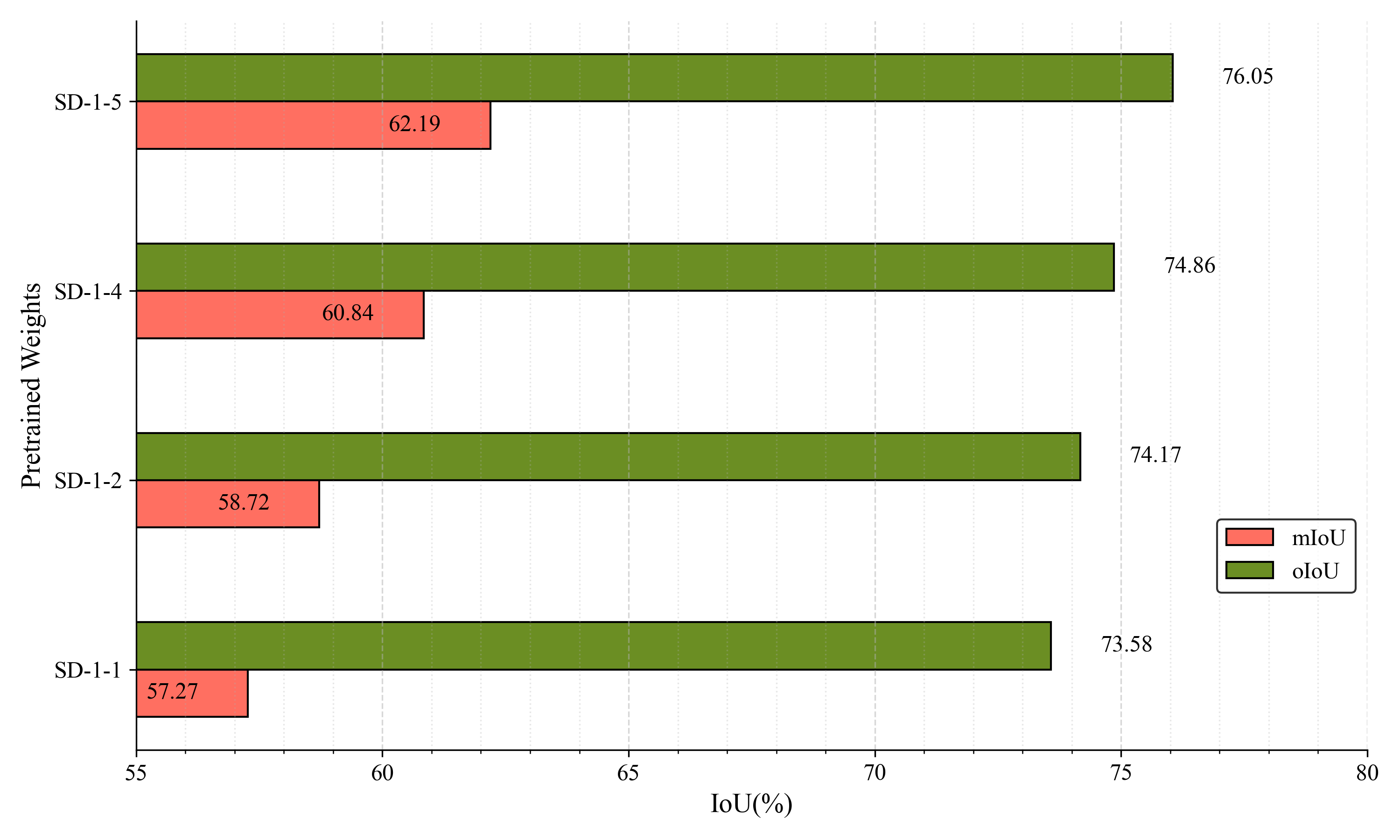}}
		
		\caption{Performance comparison of DiffRIS using different pre-trained weights from Stable Diffusion models on the RRSIS-D test set. The metrics mIoU and oIoU are plotted for four versions of Stable Diffusion (SD-1-1, SD-1-2, SD-1-4, SD-1-5), demonstrating a consistent improvement in performance with increasing pre-training iterations.}\label{ablation_weight}
	\end{center}
\end{figure}

As depicted in Fig.~\ref{ablation_weight}, a clear and consistent improvement in both mIoU and oIoU metrics is observed as the number of pre-training iterations increases. Notably, we observe a steady upward trend in performance across the various releases of the Stable Diffusion model, from SD-1-1 to SD-1-5. The mIoU value for SD-1-5 reaches 62.19\%, markedly outperforming SD-1-1, which attains only 57.27\%. This represents a significant enhancement of 4.92\% in mIoU. Similarly, the oIoU score for SD-1-5 rises to 76.05\%, surpassing the 73.58\% of SD-1-1 by 2.47\%. These results underscore a positive correlation between the number of pre-training iterations and the model's ability to capture finer details, thereby producing more accurate and robust results on the RRSIS task.

\subsubsection{Effects on the CP-adapter}

To investigate the efficacy of the CP-adapter, we conduct an ablation study on the RRSIS-D test set within the DiffRIS framework, progressively adding key components of the text adapter. As shown in Table~\ref{ablation_adapter}, when the CP-adapter is removed, DiffRIS achieves the lowest segmentation performance, with a mIoU of 75.52\% and an oIoU of 61.43\%. These results underscore the critical role of the CP-adapter in RRSIS tasks, as its absence significantly impairs the model's ability to preserve contextual information and align features effectively.

Subsequently, we incrementally reintroduce the three core components of the CP-adapter: global context modeling, object-aware reasoning, and domain-specific adjustment. The introduction of global context modeling does not yield a significant performance boost, primarily due to its limited capacity to address the nuanced, domain-specific challenges inherent in remote sensing tasks. However, as object-aware reasoning is added, the oIoU improves to 61.82\%, and with the subsequent inclusion of domain-specific adjustment, the oIoU further increases to 62.19\%. These results demonstrate the importance of each component, where global context modeling lays the foundation for broader contextual understanding, while object-aware reasoning and domain-specific adjustment are pivotal in refining the model’s RIS accuracy and domain adaptation capabilities.

Overall, the complete CP-adapter effectively bridges the domain gap between pre-training tasks and downstream remote sensing applications, significantly enhancing RRSIS performance. The CP-adapter refines linguistic features to adapt them to the specific demands of RRSIS tasks, ensuring optimal feature alignment and task adaptability. These findings highlight the indispensable role of the CP-adapter in improving the model’s RIS performance through its lightweight yet efficient architecture.

\begin{table}[t]
	\centering
	\caption{Ablation Analysis of CP-adapter with Different Components. The Best Performance Is Shown in Bold.}
	\label{ablation_adapter}
	\renewcommand\arraystretch{1.0}
	\fontsize{8}{12}\selectfont
	\begin{tabular}{c|l|c|c}
		\hline \hline
		& \textbf{Method}  & \textbf{mIoU} & \textbf{oIoU} \\ \hline
		1  & w/o CP-adapter              &  75.52  &  61.43  \\ 
		2  & 1 + global context modeling     &   75.43 & 61.49     \\ 
		3  & 2 + object-aware reasoning          &   75.88 & 61.82    \\ 
		4  & 3 + domain-specific adjustment   &    \textbf{76.05} & \textbf{62.19}    \\ \hline \hline
	\end{tabular}
\end{table}

\subsubsection{Effects on the PCMRD}

We first compare PCMRD with three other decoders: LMP, OAD, and CAMD, and the results are shown in Table~\ref{com_decoder}. It is clear that PCMRD achieves the highest precision at multiple thresholds (Pr@0.5, Pr@0.7, and Pr@0.9), significantly outperforming the competing methods. This superior performance is attributed to the novel progressive cross-modal reasoning mechanism in PCMRD, which facilitates fine-grained interactions between visual and linguistic features. Unlike the decoders in LMP and OAD, which do not leverage cross-modal feature interactions, PCMRD integrates dynamic query-text and query-vision interactions, enabling more accurate object localization and segmentation. Furthermore, while CAMD captures valuable visual features, it neglects multi-scale information, which is crucial for handling objects of varying sizes and complex scenarios. In contrast, PCMRD excels in this aspect by employing multi-scale feature fusion, leading to enhanced performance in both global and local context understanding.

\begin{table}[t]
	\centering
	\caption{Ablation Studies on Decoders in DiffRIS. The Best Performance Is Shown in Bold.}
	\label{com_decoder}
	\renewcommand\arraystretch{1.0}
	\fontsize{8}{12}\selectfont
	\begin{tabular}{l|ccc|cc}
		\hline \hline
		\textbf{Decoder} & \textbf{Pr@0.5} & \textbf{Pr@0.7} & \textbf{Pr@0.9} & \textbf{mIoU} & \textbf{oIoU}  \\ \hline
		LMP \cite{yang2022lavt}      &  63.82  & 48.65 & 20.73 & 75.98 & 56.54 \\ 
		OAD \cite{liu2024rotated}     &  65.99  & 49.83 & 21.52 & 75.62 & 58.31   \\ 
		CAMD \cite{liu2023caris}        &  71.32  & 52.17 & 23.51 & \textbf{76.41} & 61.86  \\ 
		PCMRD (ours)    & \textbf{71.79} & \textbf{54.64} & \textbf{26.46} & 76.05 & \textbf{62.19}   \\ \hline \hline
	\end{tabular}
\end{table}

\begin{table}[t]
	\centering
	\caption{Ablation Analysis of PCMRD with Different Components. The Best Performance Is Shown in Bold.}
	\label{ablation_PCMRD}
	\renewcommand\arraystretch{1.0}
	\fontsize{8}{12}\selectfont
	\begin{tabular}{c|l|c|c}
		\hline \hline
		& \textbf{Method}  & \textbf{mIoU} & \textbf{oIoU}  \\ \hline
		1  & baseline              &  72.18  &  53.62  \\ 
		2  & 1 + multi-scale fusion     &   74.29 & 56.37     \\ 
		3  & 2 + query tokens          &   75.46 & 59.81    \\ 
		4  & 3 + hard assignment   &    \textbf{76.05} & \textbf{62.19}    \\ \hline \hline
	\end{tabular}
\end{table}

Moreover, the ablation study in Table \ref{ablation_PCMRD} highlights the contribution of each component in the PCMRD. The baseline is defined as the model that directly extracts visual features and predicts the segmentation results using dynamic convolutions from the linguistic features. In the baseline, introducing multi-scale fusion improves performance by capturing both global and local contextual information, resulting in an increase of mIoU to 74.29\% and oIoU to 56.37\%. Further, adding query tokens refines object-level understanding, leading to a further improvement in RRSIS performance. Finally, the use of hard assignment with learnable Gumbel Softmax refines token grouping, achieving the best performance with an mIoU of 76.05\% and oIoU of 62.19\%. These results demonstrate the effectiveness of each component of PCMRD in enhancing segmentation accuracy and object localization.

\section{Conclusion}
\label{section:Conclusion}

This paper introduces DiffRIS, a novel framework that enhances RRSIS tasks by leveraging the rich knowledge embedded in pre-trained text-to-image diffusion models. By integrating two key modules—CP-adapter, which refines cross-modal alignment for enhanced semantic understanding, and PCMRD, which leverages multi-scale contextual information to improve segmentation accuracy—DiffRIS significantly advances the performance of RRSIS, overcoming challenges such as varying object scales and semantic ambiguities in remote sensing imagery.

Despite the promising results, DiffRIS has some limitations that require further attention. First, the model's interpretability remains limited, making it challenging to fully understand how textual descriptions are mapped to complex remote sensing features. Second, its performance in extreme conditions, such as images with adverse weather or rare geographical contexts, needs further investigation. Additionally, the computational cost of large pre-trained models may hinder real-time deployment in resource-constrained environments. Future research should focus on improving model interpretability through explainable techniques, enhancing robustness in extreme scenarios, and optimizing computational efficiency for practical, real-time applications.

\section*{Acknowledgments}

This work was supported by National Science Fund for Outstanding Young Scholars under Grant 62025107.

\bibliographystyle{elsarticle-num} 
\bibliography{refs}

\end{document}